\documentclass[11pt,a4paper]{article}
  \usepackage[hyperref]{emnlp2018}
  \usepackage{times}
  \usepackage{latexsym}
  \usepackage{url}
  \usepackage{times}
  \usepackage{xcolor}
  \usepackage{soul}
  \usepackage{times}
  \usepackage{latexsym}
  \usepackage{amsfonts}
  \usepackage{amsmath}
  \usepackage{bbm}
  \usepackage{graphicx}
  \usepackage{amssymb}
  \usepackage{subfigure}
  \usepackage{balance}
  \usepackage{threeparttable}
  \usepackage{multirow}
  \usepackage[utf8]{inputenc}
  \hypersetup{draft} 
  
  \aclfinalcopy 
  
  \title{Differentiating Concepts and Instances for Knowledge Graph Embedding}

  \newcommand*{\affaddr}[1]{#1}
  
  \newcommand*{\email}[1]{\texttt{#1}}

  \author{
    \textbf{Xin Lv}, \textbf{Lei Hou}\thanks{corresponding author}, \textbf{Juanzi Li}, \textbf{Zhiyuan Liu}\\
    \affaddr{Department of Computer Science and Technology,\\
    Tsinghua University, China 100084}\\
    \email{\{lv-x18@mails.,houlei@,lijuanzi@,liuzy@\}tsinghua.edu.cn}
    }
  
  \date{}
  
\begin{document}
  \maketitle
  \begin{abstract}
    Concepts, which represent a group of different instances sharing common properties, are essential information
    in knowledge representation. Most conventional knowledge embedding methods encode both entities (concepts and instances) and relations 
    as vectors in a low dimensional semantic space equally, ignoring the difference between concepts and instances. 
    In this paper, we propose a novel knowledge graph embedding model named TransC by differentiating concepts and 
    instances. Specifically, TransC encodes each concept in knowledge graph as a sphere and each instance as a 
    vector in the same semantic space. We use the relative positions to model the relations between concepts and 
    instances (i.e., \texttt{instanceOf}), and the relations between concepts and sub-concepts (i.e., \texttt{subClassOf}). We evaluate 
    our model on both link prediction and triple classification tasks on the dataset based on YAGO. Experimental results 
    show that TransC outperforms state-of-the-art methods, and captures the semantic transitivity for \texttt{instanceOf} 
    and \texttt{subClassOf} relation. Our codes and datasets can be obtained from \url{https://github.com/davidlvxin/TransC}.
  \end{abstract}
  
  \section{Introduction}
  
  Knowledge graphs (KGs) aim at semantically representing the world's truth in the form of machine-readable graphs 
  composed of triple facts. Knowledge graph embedding encodes each element (entities and relations) 
  in knowledge graph into a continuous low-dimensional vector space. The learned representations make the knowledge graph 
  essentially computable and have been proved to be helpful for knowledge graph completion and 
  information extraction \cite{TransE,TransH,TransR,TransD,TranSparse}.
  
  \begin{figure}[ht]
    \centering
    \setlength{\abovecaptionskip}{2pt}
    \setlength{\belowcaptionskip}{0pt}
    \includegraphics[width=75.0mm]{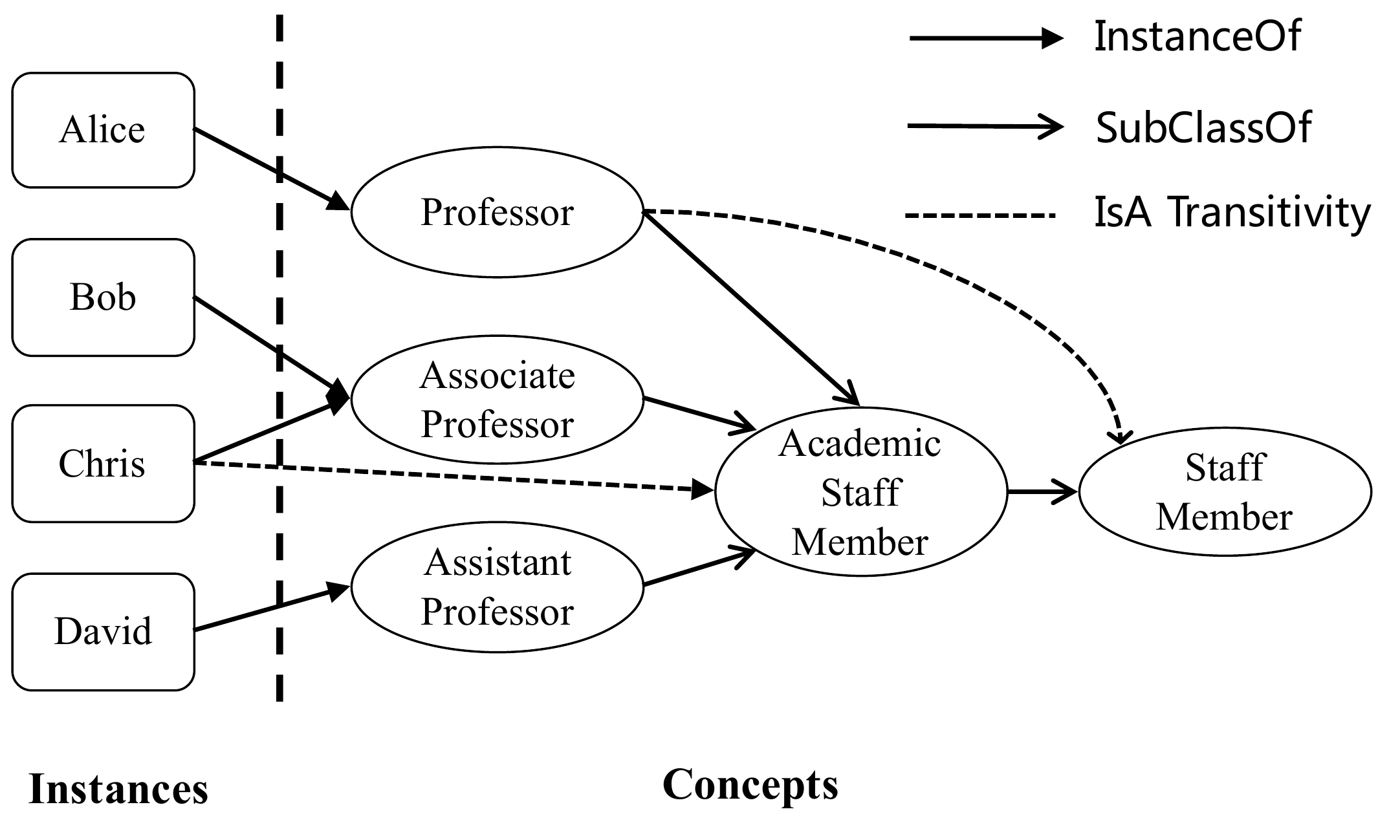}
    \caption{An example of concepts, instances, and isA transitivity.} 
    \label{fig_intro}
    \end{figure}
   
  In recent years, various knowledge graph embedding methods have been proposed, among which the translation-based models 
  are simple and effective with good performances. Inspired by word2vec \cite{Word2Vec}, given a triple 
  (h, r, t), TransE learns vector embeddings $\mathbf{h}$, $\mathbf{r}$ and $\mathbf{t}$ which satisfy  
  $\mathbf{r} \approx \mathbf{t} - \mathbf{h}$. Afterwards, 
  TransH \cite{TransH}, TransR/CTransR \cite{TransR} and TransD \cite{TransD}, etc are proposed to address 
  the problem of TransE when modeling 1-to-N, N-to-1, and N-to-N relations. 
  As extensions of RESCAL\cite{RESCAL}, which is a bilinear model, HolE\cite{HolE}, DistMult\cite{DistMult}
  and ComplEx\cite{complEx} achieve the state-of-the-art performances.
  Meanwhile, there are also some different 
  methods using a variety of external information such as entity types \cite{DKRL}, 
  textual descriptions \cite{TEKE}, as well as logical rules to strengthen representations of knowledge graphs \cite{Wang2015Knowledge,Shu2016Jointly,Rockt2015Injecting}. 
  
  However, all these methods ignore to distinguish between concepts and instances, and regard both as entities to make a simplification.
  Actually, concepts and instances are organized differently in many real world datasets like YAGO \cite{YAGO}, Freebase \cite{Freebase},
  and WordNet \cite{Wordnet}. Hierarchical concepts in these knowledge bases provide a natural way to categorize and locate instances.
  Therefore, the common simplification in previous work will lead to the following two drawbacks:
  
  \textbf{Insufficient concept representation:} Concepts are essential information in knowledge graph. A concept is a 
  fundamental category of existence \cite{rosch:natural} and can be reified by all of its actual or potential instances. Figure \ref{fig_intro} presents an example of concepts and instances about university staffs.
  Most knowledge embedding methods encode both concepts and instances as vectors, cannot explicitly represent 
  the difference between concepts and instances.
  
  \textbf{Lack transitivity of both isA relations:} \texttt{instanceOf} and \texttt{subClassOf} (generally known as isA) are two special relations in 
  knowledge graph. Different from most other relations, isA relations exhibit transitivity, e.g., the dotted lines in Figure \ref{fig_intro} represent the facts inferred by isA transitivity. 
  The indiscriminate vector representation for all relations in previous work cannot reserve this property well (see Section \ref{triple classification} for details).
  
  To address these issues, we propose a novel translation embedding model named TransC in this paper. 
  Inspired by \cite{mind}, concepts in people's mind are organized hierarchically and instances should 
  be close to concepts that they belong to.
  Hence in TransC, each concept is encoded as a sphere and each instance as a vector in the same semantic space, and relative 
  positions are employed to model the relations between concepts and instances. More specifically, \texttt{instanceOf} 
  relation is naturally represented by checking whether an instance vector is inside a concept sphere. For the 
  \texttt{subClassOf} relation, we enumerate and quantify four possible relative positions between two concept spheres. 
  We also define loss functions to measure the relative positions and optimize knowledge graph embeddings.
  Finally, we incorporate them into  translation-based models to jointly learn the knowledge
  representations of concepts, instances and relations.
  
  Experiments on real world datasets extracted 
  from YAGO show that TransC outperforms previous work like TransE, TransD, HolE, DistMult and ComplEx in most cases. 
  The contributions of this paper 
  can be summarized as follows:
  
  \begin{enumerate}
    \item To the best of our knowledge, we are the first to propose and formalize the problem of 
    knowledge graph embedding which differentiates between concepts and instances.
    \item We propose a novel knowledge embedding method named TransC, which distinguishes between concepts and instances
    and deals with the transitivity of isA relations.
    \item We construct a new dataset based on YAGO for evaluation. 
    Experiments on link prediction and triple classification demonstrate that TransC successfully addresses the above problems and 
    outperforms state-of-the-art methods.
  
  \end{enumerate}  
  
  \section{Related Work}
  
  There are a variety of models for knowledge graph embedding. We divide them into three kinds and introduce them respectively.
  
  \subsection{Translation-based Models}
  
  \textbf{TransE \cite{TransE}} regards a relation $\mathbf{r}$ as a translation from $\mathbf{h}$ to $\mathbf{t}$ for a triple
  $(h, r, t)$ in training set. The vector embeddings of this triple should satisfy $\mathbf{h} + \mathbf{r} \approx \mathbf{t}$.
  Hence, $\mathbf{t}$ should be the nearest neighbor of $\mathbf{h} + \mathbf{r}$, and the loss function is 
  \begin{equation}
    f_r(h, t) = ||\mathbf{h} + \mathbf{r} - \mathbf{t}||_2^2.
  \end{equation}
  TransE is suitable for 1-to-1 relations, but it has problems when handling 1-to-N, N-to-1, and N-to-N relations.
  
  \noindent\textbf{TransH \cite{TransH}} attempts to alleviate the problems of TransE above. It regards a relation vector $\mathbf{r}$ 
  as a translation on a hyperplane with $\mathbf{w}_r$ as the normal vector. The vector embeddings will be first projected to the 
  hyperplane of relation $r$ and get $\mathbf{h}_{\perp} = \mathbf{h} - \mathbf{w}_r^{\top}\mathbf{h}\mathbf{w}_r$
  and $\mathbf{t}_{\perp} = \mathbf{t} - \mathbf{w}_r^{\top}\mathbf{t}\mathbf{w}_r$.
  The loss function of TransH is
  \begin{equation}
    f_r(h, t) = ||\mathbf{h}_{\perp} + \mathbf{r} - \mathbf{t}_{\perp}||_2^2.
  \end{equation} 
  
  \noindent\textbf{TransR/CTransR \cite{TransR}} addresses the issue in TransE and TransH that some entities are similar in the entity space but comparably 
  different in other specific aspects. 
  It sets a transfer matrix $\mathbf{M}_r$ for each relation $r$ to map entity embedding to relation vector space. 
  Its loss function is
  \begin{equation}
    f_r(h, t) = ||\mathbf{M}_r\mathbf{h} + \mathbf{r} - \mathbf{M}_r\mathbf{t}||_2^2.
  \end{equation} 
  
  \noindent\textbf{TransD \cite{TransD}} considers the different types of entities and relations at the same time. Each 
  relation-entity pair $(r, e)$ will have a mapping matrix $\mathbf{M}_{re}$ to map entity embedding into relation vector space.
  And the projected vectors could be defined as  $\mathbf{h}_\perp = \mathbf{M}_{rh}\mathbf{h}$ and $\mathbf{t}_\perp = \mathbf{M}_{rt}\mathbf{t}$. The loss
  function of TransD is
  \begin{equation}
    f_r(h, t) = ||\mathbf{h}_{\perp} + \mathbf{r} - \mathbf{t}_{\perp}||_2^2.
  \end{equation}
  
  There are many other translation-based models in recent years. For example, 
  TranSparse \cite{TranSparse} simplifies TransR by enforcing the sparseness on the projection matrix, 
  PTransE \cite{PTransE} considers relation paths as translations between entities for representation learning,
  \cite{ManifoldE} proposes a manifold-based embedding principle (ManifoldE) for precise link prediction, 
  TransF \cite{TransF} regards relation as translation between head entity vector and tail entity vector with flexible magnitude,
  \cite{TransG} proposes a new generative model TransG, and KG2E \cite{KG2E} uses Gaussian embedding to model the data uncertainty.
  All these models can be seen in \cite{Survey}.
  
  \subsection{Bilinear Models}
  
  RESCAL\cite{RESCAL} is the first bilinear model. It associates each entity with a vector to capture its latent semantics. 
  Each relation is represented as a matrix which models pairwise interactions between latent factors.
  
  Many extensions of RESCAL have been proposed by restricting bilinear functions in recent years.
  For example, DistMult \cite{DistMult} simplifies RESCAL by restricting the matrices representing relations to diagonal matrices.
  HolE\cite{HolE} combines the expressive power of RESCAL with the efficiency and simplicity of DistMult. 
  It represents both entities and relations as vectors in $\mathbb{R}^{d}$.
  ComplEx\cite{complEx} extends DistMult by introducing complex-valued embeddings so as to better model asymmetric relations.
   
  \subsection{External Information Learning Models} 
  
  External information like textual information is significant for knowledge representation.
  TEKE \cite{TEKE} uses external context information in a text corpus to represent both entities 
  and words into a joint vector space with alignment models. DKRL \cite{DKRL} directly learns entity 
  representations from entity descriptions. \cite{Wang2015Knowledge,Shu2016Jointly,Rockt2015Injecting} use logical rules to strengthen representations of knowledge graphs.
  
  All models above do not differentiate between concepts and instances. 
  To the best of our knowledge, our proposed TransC is the first attempt which represents concepts, instances, and relations differently in the same space.
  
  \begin{figure*}[ht] 
    \centering
    \setlength{\abovecaptionskip}{2pt}
    \setlength{\belowcaptionskip}{0pt}
    \includegraphics[width=160.0mm]{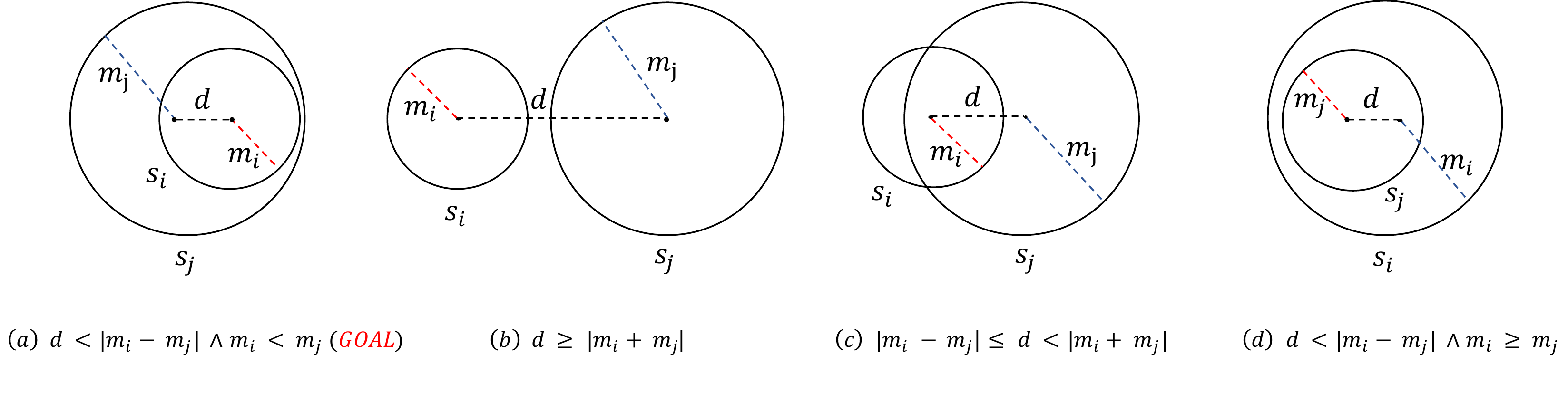}\\
    \caption{Four relative positions between sphere $s_i$ and $s_j$.}
    \label{four_position}
    \end{figure*}
   
  \section{Problem Formulation}
  
  In this section, we formulate the problem of knowledge graph embedding 
  with concepts and instances. Before that, we first introduce the input knowledge graph.
  
  \textit{\textbf{Knowledge Graph}} $\mathcal{KG}$ describes concepts, instances, and the relations between them.
  It can be formalized as $\mathcal{KG} = \{\mathcal{C}, \mathcal{I}, \mathcal{R}, \mathcal{S}\}$. $\mathcal{C}$ and $\mathcal{I}$ denote the 
  sets of concepts and instances respectively. Relation set $\mathcal{R}$ can be formalized as $\mathcal{R} = \{r_e, r_c\} \cup \mathcal{R}_l$,
  where $r_e$ is an \texttt{instanceOf} relation, $r_c$ is a \texttt{subClassOf} relation, and $\mathcal{R}_l$ is the instance relation set.
  Therefore, the triple set $\mathcal{S}$ can be divided into three disjoint subsets: 
  
  \begin{enumerate}
    \item \texttt{InstanceOf} triple set $\mathcal{S}_e = \{(i, r_e, c)_k\}_{k=1}^{n_e}$, where $i \in \mathcal{I}$ is an instance, 
    $c \in \mathcal{C}$ is a concept, and $n_e$ is the size of $\mathcal{S}_e$.
    \item \texttt{SubClassOf} triple set $\mathcal{S}_c = \{(c_{i}, r_c, c_{j})_k\}_{k=1}^{n_c}$, where $c_i, c_j \in \mathcal{C}$ 
    are concepts, $c_i$ is a sub-concept of $c_j$, and $n_c$ is the size of $\mathcal{S}_c$.
    \item Relational triple set $\mathcal{S}_l = \{(h, r, t)_k\}_{k=1}^{n_l}$, where $h, r \in \mathcal{I}$ are head instance 
    and tail instance, $r \in \mathcal{R}_l$ is an instance relation, and $n_l$ is the size of $\mathcal{S}_l$.
  \end{enumerate} 
  
  Given knowledge graph $\mathcal{KG}$, \textbf{knowledge graph 
  embedding with concepts and instances} aims at learning embeddings for instances, 
  concepts, and relations in the same space $\mathbb{R}^k$. For each concept $c \in \mathcal{C}$, 
  we learn a sphere $s(\mathbf{p}, m)$ with $\mathbf{p} \in \mathbb{R}^k$ and $m$ denoting the sphere 
  center and radius. For each instance $i \in \mathcal{I}$ and instance relation $r \in \mathcal{R}_l$,
  we learn a low-dimensional vector $\mathbf{i} \in \mathbb{R}^k$ and $\mathbf{r} \in \mathbb{R}^k$ 
  respectively. Specifically, the \texttt{instanceOf} and \texttt{subClassOf} representations 
  are well-designed so that the transitivity of isA relations can be reserved, namely,
  \texttt{instanceOf}-\texttt{subClassOf} transitivity shown in Equation \ref{formu1}:
  \begin{equation}\label{formu1}
    \resizebox{.88\hsize}{!}{$(i, r_e, c_1)\in S_e \wedge (c_1, r_c, c_2)\in S_c \rightarrow (i, r_e, c_2)\in S_e,$}
  \end{equation}
  and \texttt{subClassOf}-\texttt{subClassOf} transitivity shown in Equation \ref{formu2}:
  \begin{equation}\label{formu2}
    \resizebox{.88\hsize}{!}{$(c_1, r_c, c_2)\in S_c \wedge (c_2, r_c, c_3)\in S_c \rightarrow (c_1, r_c, c_3)\in S_c.$}
  \end{equation} 
  
  Based on the definition, how to model concepts and isA relations is critical to solve this problem.

  \section{Our Approach}
  
  To differentiate between concepts and instances for knowledge graph embedding, 
  we propose a novel method named TransC.  
  We define different loss functions to measure the relative positions in embedding space, and then
  jointly learn the representations of concepts, instances, and relations based on the 
  translation-based models.
  
  \subsection{TransC}
  
  We have three kinds of triples in our triple set $\mathcal{S}$ and define different loss function for them respectively.
  
  
  \textbf{InstanceOf Triple Representation}. For a given \texttt{instanceOf} triple $(i, r_e, c)$, 
   if it is a true triple, $\mathbf{i}$ should be inside the sphere $s$ to represent
  the \texttt{instanceOf} relation between them. 
  Actually, there is another relative position that $\mathbf{i}$ is outside the sphere $s$. In this condition, 
  the embeddings still need to be optimized. The loss function is defined as
  \begin{equation}
    f_e(i, c) = ||\mathbf{i} - \mathbf{p}||_2 - m.
  \end{equation}
  
  \textbf{SubClassOf Triple Representation}. For a \texttt{subClassOf} triple $(c_i, r_c, c_j)$, just like before, 
  concepts $c_i, c_j$ are encoded as spheres $s_i(\mathbf{p}_i, m_i)$ and $s_j(\mathbf{p}_j, m_j)$. 
  We first denote the distance between the centers of the two spheres as
  \begin{equation}
    d = ||\mathbf{p}_i - \mathbf{p}_j||_2.
  \end{equation}
  If $(c_i, r_c, c_j)$ is a true triple, sphere $s_i$ should be inside sphere $s_j$ (Figure \ref{four_position}a) to represent the \texttt{subClassOf} relation between them.
  Actually, there are three other relative positions between sphere $s_i$ and $s_j$ (as shown in Figure \ref{four_position}). We also have three loss functions under these 
  three conditions:
  \begin{enumerate}
    \item $s_i$ is separate from $s_j$ (Figure \ref{four_position}b). The embeddings still need to be optimized. In this condition, 
    the two spheres need to get closer in optimalization.
    Therefore, the loss function is defined as
    \begin{equation}
      f_c(c_i, c_j) = ||\mathbf{p}_i - \mathbf{p}_j||_2 + m_i - m_j.
    \end{equation}
    \item $s_i$ intersects with $s_j$ (Figure \ref{four_position}c). This condition is similar to condition 1. The loss function is defined as
    \begin{equation}
      f_c(c_i, c_j) = ||\mathbf{p}_i - \mathbf{p}_j||_2 + m_i - m_j.
    \end{equation}
    \item $s_j$ is inside $s_i$ (Figure \ref{four_position}d). It is different from our target and we should reduce $m_j$ and increase $m_i$. Hence, the loss function is
    \begin{equation}
      f_c(c_i, c_j) = m_i - m_j.
    \end{equation}
  \end{enumerate}
  
  \textbf{Relational Triple Representation}. For a relational triple $(h, r, t)$, TransC will learn low-dimensional 
  vectors $\mathbf{h}, \mathbf{t}, \mathbf{r} \in \mathbb{R}^k$ for instances and relations. Just like TransE \cite{TransE},
  the loss function of this kind of triples is defined as
  \begin{equation}
    f_r(h, t) = ||\mathbf{h} + \mathbf{r} - \mathbf{t}||_2^2.
  \end{equation}
  
  After having embeddings above, TransC can easily deal with the transitivity of isA relations. If we have true triples $(i, r_e, c_i)$
  and $(c_i, r_c, c_j)$, which means $\mathbf{i}$ is inside the sphere $s_i$ and $s_i$ is inside $s_j$, we can get a result that
  $\mathbf{i}$ is also inside the sphere $s_j$. It can be concluded that $(i, r_e, c_j)$ is a true triple and TransC can handle
  \texttt{instanceOf}-\texttt{subClassOf} transitivity. Similarly, if we have true triples $(c_i, r_c, c_j)$ and $(c_j, r_c, c_k)$, we can get a
  result that sphere $s_i$ is inside sphere $s_k$. It means $(c_i, r_e, c_k)$ is a true triple and TransC can deal with
  \texttt{subClassOf}-\texttt{subClassOf} transitivity.
  
  In experiments, we enforce constrains as $||\mathbf{h}||_2 \le 1$, $||\mathbf{r}||_2 \le 1$, $||\mathbf{t}||_2 \le 1$ 
  and $||\mathbf{p}||_2 \le 1$.
  
  \subsection{Training Method}
  
  For \texttt{instanceOf} triples, we use $\xi$ and $\xi'$ to denote a
  positive triple and a negative triple. $\mathcal{S}_e$ and $\mathcal{S}_e'$ are used to describe the positive triple set and 
  negative triple set. Then we can define a margin-based ranking loss for \texttt{instanceOf} triples:
  \begin{equation}
    \mathcal{L}_e = \sum_{\xi \in \mathcal{S}_e} \sum_{\xi' \in \mathcal{S}_e'}[\gamma_e + f_e(\xi) - f_e(\xi')]_{+},
  \end{equation}
  where $[x]_{+} \triangleq$ max $(0, x)$ and $\gamma_e$ is the margin separating positive triplets and negative triplets. 
  Similarly, for \texttt{subClassOf} triples, we will have a ranking loss:
  \begin{equation}
    \mathcal{L}_c = \sum_{\xi \in \mathcal{S}_c} \sum_{\xi' \in \mathcal{S}_c'}[\gamma_c + f_c(\xi) - f_c(\xi')]_{+},
  \end{equation}
  and for relational triples, we will have a ranking loss:
  \begin{equation}
    \mathcal{L}_l = \sum_{\xi \in \mathcal{S}_l} \sum_{\xi' \in \mathcal{S}_l'}[\gamma_l + f_r(\xi) - f_r(\xi')]_{+}.
  \end{equation}
  Finally, we define the overall loss function as linear combinations of these three functions:
  \begin{equation}
    \mathcal{L} = \mathcal{L}_e + \mathcal{L}_c + \mathcal{L}_l.
  \end{equation}
  The goal of training TransC is to minimize the
  above function, and iteratively update embeddings of concepts, instances, and concepts.
  
  Every triple in our training set has a label to indicate whether the triple is positive  or
  negative. But existing knowledge graph only contains positive triples. We need to generate negative triples by
  corrupting positive triples. For a relational triple $(h, r, t)$, we replace $h$ or $t$ to generate a negative triple
  $(h', r, t)$ or $(h, r, t')$. For example, we get $h'$ by randomly picking from a set 
  $\mathcal{M}_t = \mathcal{M}_1 \cup \mathcal{M}_2 \cup \dots \cup \mathcal{M}_n$, where $n$ is the number of concepts that $t$
  belongs to and $\mathcal{M}_i = \{a | a \in \mathcal{I} \land (a, r_e, c_i) \in \mathcal{S}_e \land (t, r_e, c_i) \in \mathcal{S}_e \land t \ne a \}$.
  For the other two kinds of triples, we follow the same policy to generate negative triples. We also use
  two strategies ``unif" and ``bern" described in \cite{TransH} to replace instances or concepts.
  
  \section{Experiments and Analysis}
  
  We evaluate our method on two typical tasks commonly used in knowledge graph embedding: link prediction \cite{TransE} and triple classification
  \cite{NTN}. 
  
  \subsection{Datasets}
  
  Most previous work used FB15K and WN18 \cite{TransE} for evaluation.
  But these two datasets are not suitable for our model because FB15K mainly consists of instances and WN18 mainly contains concepts.
  Therefore, we use another popular knowledge graph YAGO \cite{YAGO} for evaluation, which contains a lot of concepts from WordNet and instances from Wikipedia. 
  We construct a subset of YAGO named YAGO39K for evaluation through the following steps:
  
  \begin{table}
    \small
    \centering
    \setlength\tabcolsep{2pt}
  \begin{tabular}{|c|r|r|}
    \hline
    {DataSets} & {YAGO39K} & {M-YAGO39K}\\\hline
    \#Instance & {39,374} & {39,374}\\ 
    \#Concept & {46,110} & {46,110}\\ 
    \#Relation & {39} & {39}\\ 
    \#Relational Triple & {354,997} & {354,997}\\ 
    \#\texttt{InstanceOf} Triple & {442,836} & {442,836}\\ 
    \#\texttt{SubClassOf} Triple & {30,181} & {30,181}\\\hline
    \#Valid (Relational Triple) & {9,341} & {9,341}\\ 
    \#Test (Relational Triple) & {9,364} & {9,364}\\ 
    \#Valid (\texttt{InstanceOf} Triple) & {5,000} & {8,650}\\ 
    \#Test (\texttt{InstanceOf} Triple) & {5,000} & {8,650}\\ 
    \#Valid (\texttt{SubClassOf} Triple) & {1,000} & {1,187}\\
    \#Test (\texttt{SubClassOf} Triple) & {1,000} & {1,187}\\
    \hline 
    \end{tabular}
    \caption{\label{table1}Statistics of YAGO39K and M-YAGO39K.}
    \end{table}
    
    \begin{table*}[!htb]
      \small
      \centering
      \setlength{\belowcaptionskip}{-1pt}
          \begin{tabular}{c|cc|ccc|cccc}
          \hline
          Experiments & \multicolumn{5}{c|}{Link Prediction} & \multicolumn{4}{c}{Triple Classification(\%)} \\
          \hline
          \multirow{2}{*}{Metric} & \multicolumn{2}{c|}{MRR} & \multicolumn{3}{c|}{Hits@N(\%)} 
          & \multirow{2}{*}{Accuracy} & \multirow{2}{*}{Precision} & \multirow{2}{*}{Recall} & \multirow{2}{*}{F1-Score} \\
          & \texttt{Raw} & \texttt{Filter} & \texttt{1} & \texttt{3} & \texttt{10} &  &  &  &   \\
          \hline
          TransE  & 0.114  & 0.248  & 12.3   & 28.7  & 51.1 & 92.1  & 92.8  & 91.2   & 92.0 \\
          TransH   & 0.102  & 0.215   & 10.4   & 24.0  & 45.1 & 90.8  & 91.2   & 90.3   & 90.8  \\
          TransR   & 0.112  & 0.289   & 15.8   & 33.8 & 56.7 & 91.7  & 91.6   & 91.9   & 91.7  \\
          TransD   & 0.113  & 0.176   & 8.9   & 19.0 & 35.4 & 89.3  & 88.1   & 91.0   & 89.5  \\
          HolE & 0.063  & 0.198   & 11.0   & 23.0 & 38.4 & 92.3  & 92.6   & 91.9   & 92.3  \\
          DistMult & \textbf{0.156}  & 0.362   & 22.1   & 43.6 & 66.0 & 93.5  & 93.9   & 93.0   & 93.5  \\
          ComplEx & 0.058  & 0.362   & 29.2   & 40.7 & 48.1 & 92.8  & 92.6   & \textbf{93.1}   & 92.9  \\
          \hline
          TransC (unif)  & 0.087  & \textbf{0.421}   & 28.3   & 50.0 & 69.2 & 93.5  & 94.3   & 92.6   & 93.4  \\
          TransC (bern)  & 0.112  & 0.420   & \textbf{29.8}   & \textbf{50.2} &  \textbf{69.8} & \textbf{93.8}  & \textbf{94.8}   & 92.7   & \textbf{93.7}   \\
          \hline
          \end{tabular}
          \caption{\label{table2}Experimental results on link prediction and triple classification for relational triples. 
      Hits@N uses results of ``Filter" evaluation setting.}
      \end{table*}
  
  (1) We randomly select some relational triples like $(h, r, t)$ from the whole YAGO dataset as our relational triple set $\mathcal{S}_l$.
  
  (2) For every instance and instance relation existed in our relational triples, we save it to construct 
  instance set $\mathcal{I}$ and instance relation set $\mathcal{R}_l$ respectively.
  
  (3) For every \texttt{instanceOf} triple $(i, r_e, c)$ in YAGO, if $i \in \mathcal{I}$, we save this triple 
  to construct \texttt{instanceOf} triple set $\mathcal{S}_e$.
  
  (4) For every concept existed in \texttt{instanceOf} triple set $\mathcal{S}_e$, we save it to construct 
  concept set $\mathcal{C}$. 
  
  (5) For every \texttt{subClassOf} triple $(c_i, r_c, c_j)$ in YAGO, if $c_i \in \mathcal{C} \land c_j \in \mathcal{C}$, 
  we save this triple to construct \texttt{subClassOf} triple set $\mathcal{S}_c$.
  
  (6) Finally, we achieve our triple set $\mathcal{S} = \mathcal{S}_e \cup \mathcal{S}_c \cup \mathcal{S}_l$
  and our relation set $\mathcal{R} = \{r_e, r_c\} \cup \mathcal{R}_l$.
  
  To evaluate every model's performance in handling the transitivity of isA relations, we generate some new triples
  based on YAGO39K using the transitivity of isA relations. These new triples will be added to
  valid and test datasets of YAGO39K to create a new dataset named M-YAGO39K. Specific steps are described as follows:
  
  (1) For every \texttt{instanceOf} triple $(i, r_e, c)$ in valid and test dataset, if $(c, r_c, c_j)$ exists in training dataset, 
  we save a new \texttt{instanceOf} triple $(i, r_e, c_j)$.
  
  (2) For every \texttt{subClassOf} triple $(c_i, r_c, c_j)$ in valid and test dataset, if $(c_j, r_c, c_k)$ exists in training dataset, 
  we save a new \texttt{subClassOf} triple $(c_i, r_c, c_k)$.
  
  (3) We add these new triples to valid and test dataset of YAGO39K to get M-YAGO39K.
  
  The statistics of YAGO39K and M-YAGO39K are shown in Table \ref{table1}.

  \subsection{Link Prediction}
  
  Link Prediction aims to predict the missing $h$ or $t$ for a relational triple $(h, r, t)$. In this task, we need to
  give a ranking list of candidate instances from the knowledge graph, instead of only giving one best result.
  
  For every test relational triple $(h, r, t)$, we remove the head or tail instance and replace it
  with all instances existed in knowledge graph, and rank these instances in ascending order of distances calculated by 
  loss function $f_r$. Just like \cite{TransE}, we use two evaluation metrics in this task: (1) the mean reciprocal rank
  of all correct instances (MRR) and (2) the proportion of correct instances that rank no larger than N (Hits@N).
  A good embedding model should achieve a high MRR and a high Hits@N. We note that a corrupted triple may 
  also exist in knowledge graph, which should also be regarded as a correct prediction. However, the above evaluations 
  do not handle this issue and may underestimate the results. Hence, we filter out every triple appeared in our knowledge
  graph before getting the ranking list. The first evaluation setting is called ``Raw" and the second one is called ``Filter."
  We report the experiment results on both settings.
  
  In this task, we use dataset YAGO39K for evaluation. We select 
  learning rate $\lambda$ for SGD among \{0.1, 0.01, 0.001\}, the three margins $\gamma_l$, $\gamma_e$ and $\gamma_c$
  among \{0.1, 0.3, 0.5, 1, 2\}, the dimension of instance vectors and relation vectors $n$ among \{20, 50, 100\}.
  The best configurations are determined according to the Hits@10 in valid set. The optimal configurations are: 
  $\gamma_l = 1$, $\gamma_e = 0.1$, $\gamma_c = 1$, $\lambda = 0.001$, $n = 100$ and taking $L_2$ as dissimilarity.
  We train every model for 1000 rounds in this task.
  
  Evaluation results on YAGO39K are shown in Table \ref{table2}. From the table, we can conclude that: (1) TransC significantly outperforms other
  models in terms of Hits@N. This indicates
  that TransC can use isA triples' information better than other models, which is helpful for instance representation learning.
  (2) TransC performs a little bit worse than DistMult in some settings. 
  The reason may be that we determine the 
  best configurations only according to the Hits@10, which may lead to a low MRR. (3) The ``bern" sampling trick works well for 
  TransC.
  
  \begin{table*}[!htb]
    \centering
    \setlength{\belowcaptionskip}{-1pt}
    \small
        \begin{tabular}{c|cccc|cccc}
        \hline
        Datasets & \multicolumn{4}{c|}{YAGO39K} & \multicolumn{4}{c}{M-YAGO39K}\\
        \hline
        \multirow{1}{*}{Metric} & Accuracy & Precision & Recall & F1-Score & Accuracy & Precision & Recall & F1-Score \\
        \hline
        TransE  & 82.6  & 83.6  & 81.0   & 82.3 & 71.0$\downarrow$   & 81.4$\downarrow$  & 54.4$\downarrow$   & 65.2$\downarrow$   \\
        TransH  & 82.9  & 83.7   & 81.7   & 82.7 & 70.1$\downarrow$  & 80.4$\downarrow$   & 53.2$\downarrow$   & 64.0$\downarrow$  \\
        TransR  & 80.6  & 79.4   & \textbf{82.5}   & 80.9  & 70.9$\downarrow$  & 73.0$\downarrow$   & 66.3$\downarrow$   & 69.5$\downarrow$ \\
        TransD  & 83.2  & 84.4   & 81.5   & 82.9   & 72.5$\downarrow$  & 73.1$\downarrow$   & 71.4$\downarrow$   & 72.2$\downarrow$\\
        HolE    & 82.3  & 86.3   & 76.7   & 81.2 & 74.2$\downarrow$  & 81.4$\downarrow$   & 62.7$\downarrow$   & 70.9$\downarrow$  \\
        DistMult & \textbf{83.9}  & \textbf{86.8}   & 80.1   & \textbf{83.3} & 70.5$\downarrow$  & 86.1$\downarrow$   & 49.0$\downarrow$   & 62.4$\downarrow$  \\
        ComplEx & 83.3  & 84.8   & 81.1   & 82.9 & 70.2$\downarrow$  & 84.4$\downarrow$   & 49.5$\downarrow$   & 62.4$\downarrow$  \\
        \hline
        TransC (unif)  & 80.2  & 81.6   & 80.0   & 79.7  & \textbf{85.5}$\uparrow$  & \textbf{88.3}$\uparrow$   & 81.8$\uparrow$   & 85.0$\uparrow$ \\
        TransC (bern)  & 79.7  & 83.2   & 74.4   & 78.6 & 85.3$\uparrow$  & 86.1$\uparrow$   & \textbf{84.2}$\uparrow$   & \textbf{85.2}$\uparrow$ \\
        \hline
        \end{tabular}
        \caption{\label{table3}Experimental results on \texttt{instanceOf} triple classification(\%).}
    \end{table*}
  
  \begin{table*}[!htb]
    \centering
    \setlength{\belowcaptionskip}{-1pt}
    \small
        \begin{tabular}{c|cccc|cccc}
        \hline
        Datasets & \multicolumn{4}{c|}{YAGO39K} & \multicolumn{4}{c}{M-YAGO39K}\\
        \hline
        \multirow{1}{*}{Metric} & Accuracy & Precision & Recall & F1-Score & Accuracy & Precision & Recall & F1-Score \\
        \hline
        TransE  & 77.6  & 72.2  & 89.8   & 80.0 & 76.9$\downarrow$  & 72.3$\uparrow$  & 87.2$\downarrow$   & 79.0$\downarrow$   \\
        TransH  & 80.2  & 76.4   & 87.5   & 81.5  & 79.1$\downarrow$  & 72.8$\downarrow$   & 92.9$\uparrow$   & 81.6$\uparrow$ \\
        TransR  & 80.4  & 74.7   & 91.9   & 82.4  & 80.0$\downarrow$  & 73.9$\downarrow$   & 92.9$\uparrow$   & 82.3$\downarrow$ \\
        TransD  & 75.9  & 70.6   & 88.8   & 78.7   & 76.1$\uparrow$  & 70.7$\uparrow$   & 89.0$\uparrow$   & 78.8$\uparrow$\\
        HolE    & 70.5  & 73.9   & 63.3   & 68.2   & 66.6$\downarrow$  & 72.3$\downarrow$   & 53.7$\downarrow$   & 61.7$\downarrow$\\
        DistMult & 61.9  & 68.7   & 43.7   & 53.4   & 60.7$\downarrow$  & 71.7$\uparrow$   & 35.5$\downarrow$   & 47.7$\downarrow$\\
        ComplEx & 61.6  & 71.5   & 38.6   & 50.1   & 59.8$\downarrow$  & 65.6$\downarrow$   & 41.4$\uparrow$   & 50.7$\uparrow$\\
        \hline
        TransC (unif)  & 82.9  & 77.1   & 93.7   & 84.6  & 83.0$\uparrow$  & 77.5$\uparrow$   & \textbf{93.1}$\downarrow$   & 84.7$\uparrow$ \\
        TransC (bern)  & \textbf{83.7}  & \textbf{78.1}   & \textbf{93.9}   & \textbf{85.2}  & \textbf{84.4}$\uparrow$  & \textbf{80.7}$\uparrow$   & 90.4$\downarrow$  & \textbf{85.3}$\uparrow$ \\
        \hline
        \end{tabular} 
        \caption{\label{table4}Experimental results on \texttt{subClassOf} triple classification(\%).}
    \end{table*}
  
  \subsection{Triple Classification}\label{triple classification}
  
  Triple Classification aims to judge whether a given triple is correct or not, which is a binary classification task. 
  This triple can be a relational triple, an \texttt{instanceOf} triple or a \texttt{subClassOf} triple. 
  
  Negative triples are needed for evaluation of binary classification. Hence, we construct some negative triples 
  following the same setting in \cite{NTN}. There are as many true triples as negative triples in 
  both valid and test set.
  
  For triple classification, we set a threshold $\delta_r$ for every relation $r$. For a given test triple, 
  if its loss function is smaller than $\delta_r$, it will be classified as positive, otherwise negative. $\delta_r$
  is obtained by maximizing the classification accuracy on valid set.
  
  In this task, we use dataset YAGO39K and M-YAGO39K for evaluation.  
  Parameters are selected in the same way as in link prediction.
  The best configurations are determined by accuracy in valid set. The optimal configurations for YAGO39K are:
  $\gamma_l = 1$, $\gamma_e = 0.1$, $\gamma_c = 0.1$, $\lambda = 0.001$, $n = 100$ and taking $L_2$ as dissimilarity.
  The optimal configurations for M-YAGO39K are: $\gamma_l = 1$, $\gamma_e = 0.1$, $\gamma_c = 0.3$, $\lambda = 0.001$, 
  $n = 100$ and taking $L_2$ as dissimilarity. For both datasets, we traverse all the training triples for 1000 rounds.
  
  Our datasets have three kinds of triples. Hence, we do experiments on them respectively. Experimental results
  for relational triples, \texttt{instanceOf} triples, and \texttt{subClassOf} triples are shown in Table \ref{table2}, Table \ref{table3}, and Table \ref{table4} respectively.
  In Table \ref{table3} and Table \ref{table4}, a rising arrow means performance of this model have a promotion from YAGO39K to M-YAGO39K and a down arrow
  means a drop.
  
  From Table \ref{table2}, we can learn that: (1) TransC outperforms all previous work 
  in relational triple classification. (2) The ``bern" sampling trick works better than ``unif" in TransC.
  
  From Table \ref{table3} and Table \ref{table4}, we can conclude that: (1) On YAGO39K,  some compared models perform better than TransC in \texttt{instanceOf}
  triple classification. This is because that \texttt{instanceOf} has most triples (53.5\%) among all relations in YAGO39K. 
  This relation is trained superabundant times and nearly achieves the best performance, which has an adverse effect on
  other triples. TransC can find a balance between them and all triples achieve a good performance.
  (2) On YAGO39K, TransC outperforms other models in \texttt{subClassOf} triple classification. As shown in Table \ref{table1}, \texttt{subClassOf} triples are much 
  less than \texttt{instanceOf} triples. Hence, other models can not achieve the best performance under 
  the bad influence of \texttt{instanceOf} triples. (3) On M-YAGO39K, TransC outperforms previous work in both \texttt{instanceOf} triple
  classification and \texttt{subClassOf} triple classification, which indicates that TransC can handle the transitivity of isA relations
  much better than other models. (4) After comparing experimental results in YAGO39K and M-YAGO39K, we can find that most previous work's performance 
  suffers a big drop in \texttt{instanceOf} triple classification and a small drop in \texttt{subClassOf} triple classification.
  This shows that previous work can not deal with \texttt{instanceOf}-\texttt{subClassOf} transitivity well.
  (5) In TransC, nearly all performances have a significant promotion from YAGO39K to M-YAGO39K. Both \texttt{instanceOf}-\texttt{subClassOf} transitivity and
  \texttt{subClassOf}-\texttt{subClassOf} transitivity are solved well in TransC.
  
  \subsection{Case Study}
  
  We have shown that TransC have a good performance for knowledge graph embedding and dealing with transitivity of isA relations.
  In this section, we present an example of finding new \texttt{instanceOf} triples and \texttt{subClassOf} triples using results of TransC.
  
\begin{figure}[ht]
    \centering
    \setlength{\abovecaptionskip}{2pt}
    \setlength{\belowcaptionskip}{0pt}
    \includegraphics[width=75.0mm]{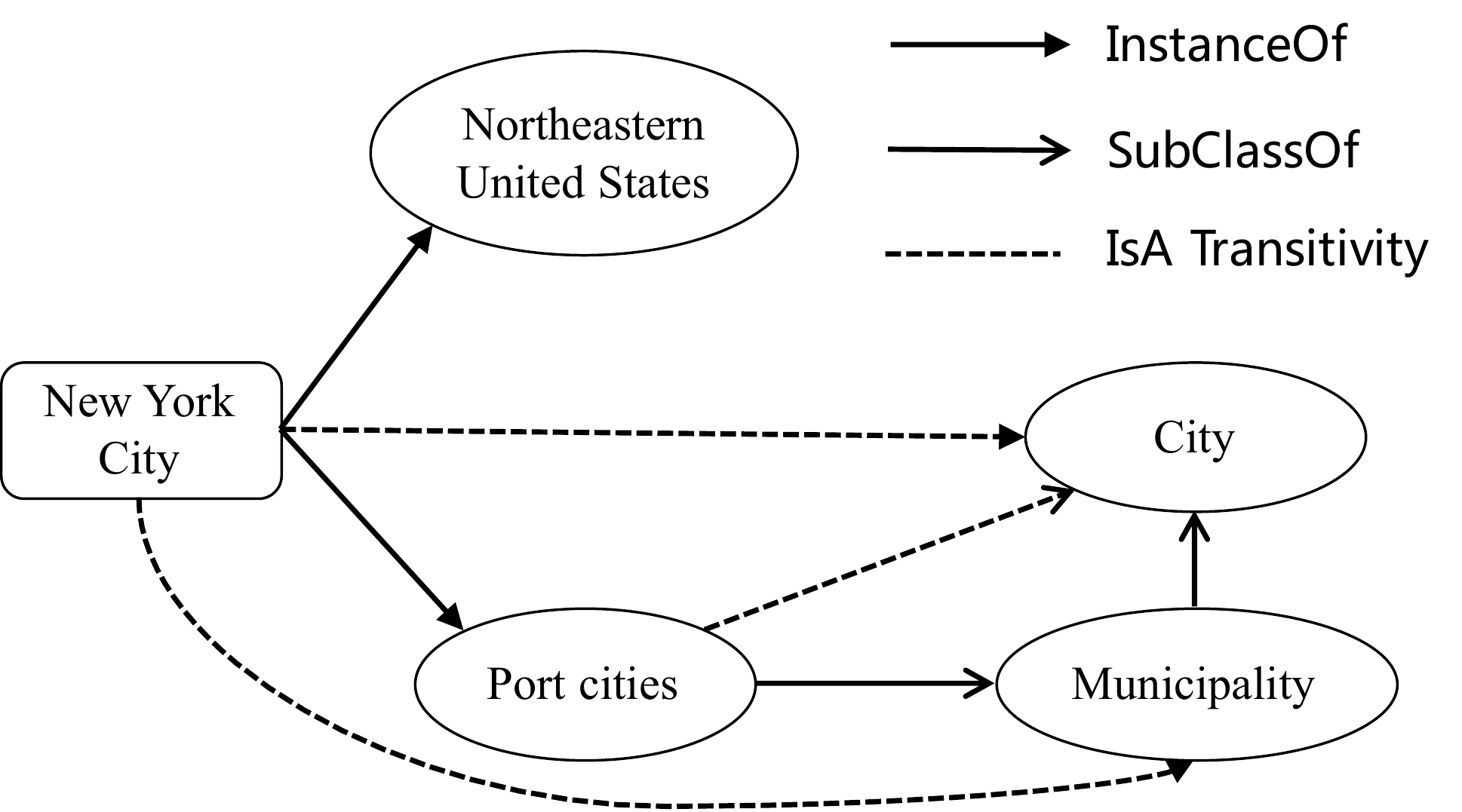}\\
    \caption{An inference example of TransC.}
    \label{fig3}
    \end{figure}

  As shown in Figure \ref{fig3}, \textit{New York City} is an instance and others are concepts. 
  The solid lines represent the triples from our datasets and the dotted lines represent the facts 
  inferred by our model. TransC can find two new \texttt{instanceOf} triples (\textit{New York City}, \texttt{instanceOf}, \textit{City}) and
  (\textit{New York City}, \texttt{instanceOf}, \textit{Municipality}). It can also find a new \texttt{subClassOf} triple (\textit{Port Cities}, \texttt{subClassOf}, \textit{City}).
  Following the transitivity of isA relations, we can know all these three new triples are right.
  Unfortunately, most previous work regards these three triples as wrong, which means they can not handle transitivity of isA relations well.
  
  \section{Conclusion and Future Work}
  
  In this paper, we propose a new knowledge embedding model named TransC. TransC embeds instances, concepts, and relations in the same space
  to deal with the transitivity of isA relations. We create a new dataset YAGO39K for evaluation. Experiment results show that 
  TransC outperforms previous translation-based models in most cases. Besides, It can also handle the transitivity of isA relations
  much better than other models. In our future work, we will explore the following research directions:
  (1) Sphere is a simple model to represent a concept in semantic space, but it still have some limits since it is too naive.
  we will try to find a more expressive model instead of spheres to represent concepts. (2) A concept may have 
  different meanings in different triples. We will try to use several typical vectors of instances as a concept's centers
  to represent different meanings of a concept. Then a concept can have different embeddings in different triples.
  
  \section*{Acknowledgments}
  
  The work is supported by NSFC key project (No. 61533018, U1736204, 61661146007), Ministry of Education and China Mobile Research Fund (No. 20181770250), and THUNUS NExT Co-Lab.
  
  \bibliography{emnlp2018}
  \bibliographystyle{acl_natbib_nourl}
  
  \appendix
  
  \end{document}